\pdfoutput=1

\documentclass[11pt]{article}

\usepackage[final]{acl}

\usepackage{times}
\usepackage{latexsym}

\usepackage[T1]{fontenc}

\usepackage[utf8]{inputenc}

\usepackage{microtype}

\usepackage{inconsolata}

\usepackage{graphicx}
\usepackage{caption}
\usepackage{subcaption}
\usepackage{xurl}
\usepackage{tabularray}
\usepackage{twemojis}
\usepackage{multirow}
\usepackage{tabularx}
\usepackage{csquotes}
\newcommand{\yes}{\scalebox{1.25}{\twemoji{check mark button}}}
\newcommand{\no}{\scalebox{1.25}{\twemoji{cross mark}}}
\newcommand{\grape}{\scalebox{1.25}{\twemoji{grapes}}}

\newcommand{\github}{\includegraphics[width=12px]{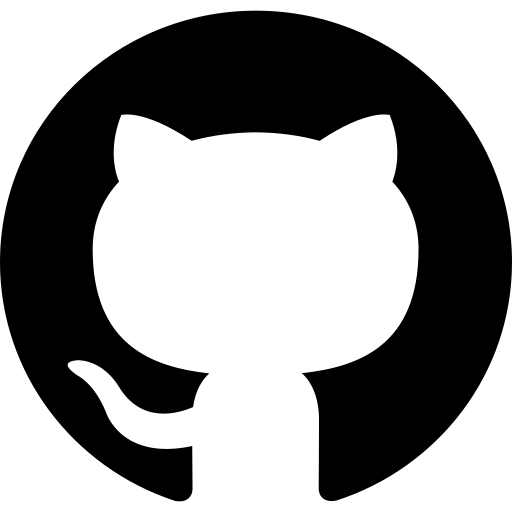}}
\newcommand{\huggingface}{\includegraphics[width=12px]{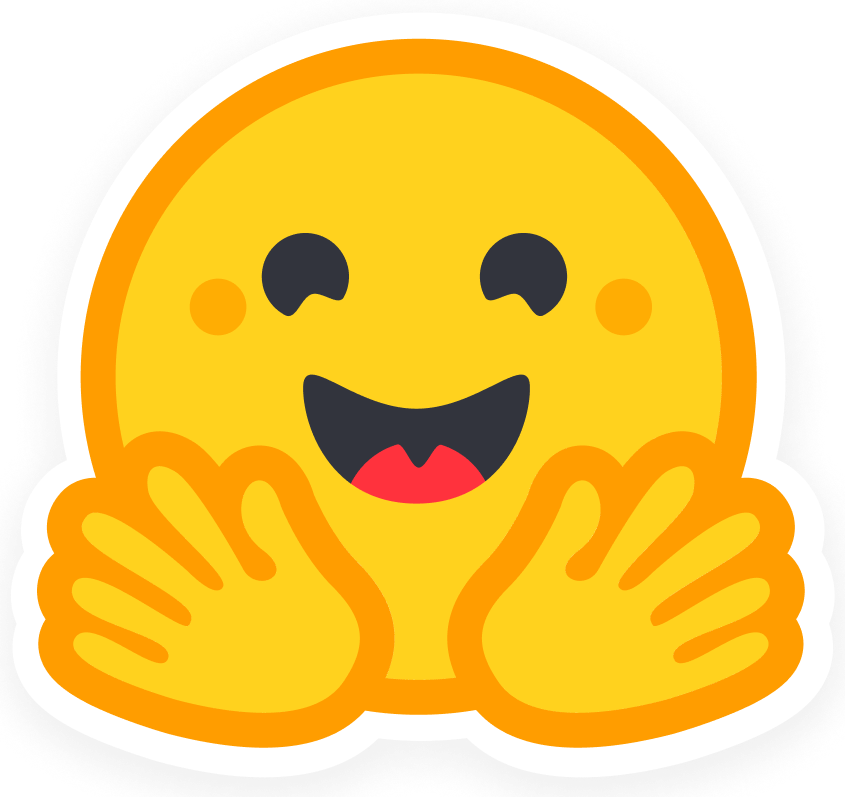}}
\newcommand{\mosel}{\includegraphics[width=80px]{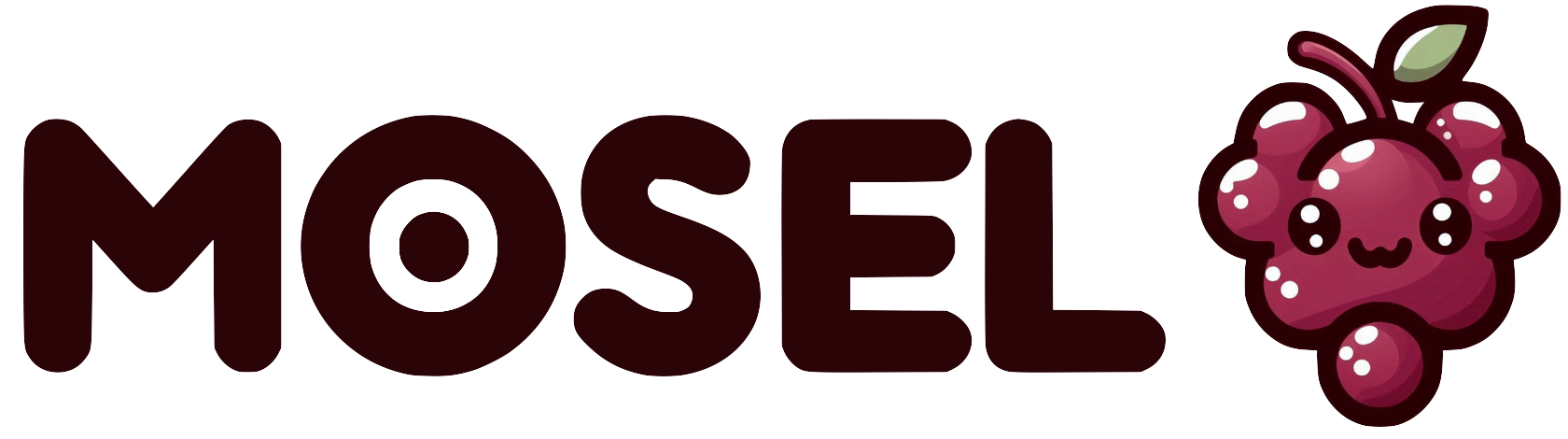}}
\newcommand{\mosels}{\textbf{MOSEL}\grape{}}

\newcommand{\mg}[1]{\textcolor{black}{#1}}
\newcommand{\mmg}[1]{\textcolor{black}{#1}}
\newcommand{\sara}[1]{\textcolor{black}{#1}}

\newcommand{\lb}[1]{\textcolor{black}{#1}}

\title{\raisebox{-.2\height}\mosel{}: 950,000 Hours of Speech Data for Open-Source \\ Speech Foundation Model Training on EU Languages}

\author{Marco Gaido\textsuperscript{\grape}, Sara Papi\textsuperscript{\grape}, Luisa Bentivogli, Alessio Brutti, Mauro Cettolo, \\\textbf{Roberto Gretter, Marco Matassoni, Mohamed Nabih, Matteo Negri} \\
  Fondazione Bruno Kessler, Italy \\
  \texttt{\{mgaido,spapi,bentivo,brutti,cettolo,gretter,matasso,mnabih,negri\}@fbk.eu} }

\begin{document}
\maketitle
\begin{abstract}
The rise of foundation models (FMs), coupled with regulatory efforts addressing their risks and impacts, has sparked significant interest in open-source models. However, existing speech FMs (SFMs) fall short of full compliance with the open-source principles, even if claimed otherwise, as no existing SFM has model weights, code, and training data publicly available under open-source terms.
In this work, we take the first step toward filling this gap by focusing on the 24 official languages of the European Union (EU). We collect suitable training data by surveying automatic speech recognition datasets and unlabeled speech corpora under open-source compliant licenses, for a total of 950k hours. Additionally, we release automatic transcripts for 441k hours of unlabeled data under the permissive CC-BY license, thereby facilitating the creation of open-source SFMs for the EU languages.
\begin{itemize}
    \item[\raisebox{-.2\height}\github{}] \href{https://github.com/hlt-mt/mosel}{\texttt{github.com/hlt-mt/mosel}}
    \item[\raisebox{-.2\height}\huggingface{}] \href{https://huggingface.co/datasets/FBK-MT/mosel}{\texttt{hf.co/datasets/FBK-MT/mosel}}
\end{itemize}
\end{abstract}

\newcommand\blfootnote[1]{%
  \begingroup
  \renewcommand\thefootnote{}\footnote{#1}%
  \addtocounter{footnote}{-1}%
  \endgroup
}
\blfootnote{\grape{} Equal contribution.}

\section{Introduction}

The introduction of foundation models trained on large datasets is revolutionizing the landscape of many NLP fields \citep{Bommasani2021FoundationModels}, particularly with the release of Large Language Models (LLMs) that demonstrated impressive abilities on various tasks \citep{radford2019language}. 
The interest attracted by such models has come together with concerns about their risks and impact, as well as requests for a better understanding of their inner workings. On the one hand, this has led to regulatory efforts \cite{ai_act,huw-etal-2024}, while on the other hand, it has sparked a growing interest in open-source models \citep{workshop2023bloom,groeneveld2024olmo} that can be accessed and studied by anyone.
However, it has been acknowledged that the term ``\textit{open source}'' has been abused \citep{eiras2024risks,Liesenfeld-2024}, \sara{being} associated with any model whose weights are free to access \citep[e.g.,][]{touvron2023llama,vicuna2023}, which is not sufficient to define a model as open source (OS).

In line with the Open Source Definition and its principles,\footnote{\url{https://opensource.org/osd}} the Open Source Initiative defines as Open Source AI a ``\textit{system made available under terms that grant the freedoms to: \textbf{use} the system for any purpose without having to ask for permission}'', ``\textit{\textbf{study}}'', ``\textit{\textbf{modify} [...] for any purpose}'', and ``\textit{\textbf{share} [...] with or without modifications, for any purpose}''.\footnote{\url{https://opensource.org/deepdive/drafts/the-open-source-ai-definition-draft-v-0-0-8}} Specifically, it requires that the model and the code \enquote{\textit{used to train and run the system}} are available under an OS license,\footnote{\url{https://opensource.org/licenses}} and that the training data is available under an OS-compliant license \citep{white2024model}.
This means that an OS model should not be trained on data released under licenses that restrict any of the four essential rights -- use, study, modify, and share -- for any purpose, including commercial use. Examples of OS-compliant licenses include CDLA-Permissive-2.0 and CC-BY-4.0, which only requires attribution (i.e., acknowledging the source or resource used). Instead, data released under licenses like CC-NC-4.0, which prohibits commercial use, or CC-SA-4.0, which mandates that derivative works have to be distributed under the same terms (thereby limiting the freedom to modify and share for any purpose), are not OS compliant.

Focusing on speech foundation models (SFMs), none of the existing ones comply with this definition.
For instance, SeamlessM4T's model \citep{communication2023seamlessm4t} is released under a license that is not OS compliant, while Whisper's model and inference code \citep{radford2023robust} are, but the training code and data are not public. 
Lastly, OWSM \citep{peng2023owsm}, although \sara{fulfilling} most of the requirements, \sara{has been trained using datasets such as MuST-C \citep{di-gangi-etal-2019-must} and SPGISpeech \citep{oneill21_interspeech}, 
which have licenses that 
do not permit derivative works or commercial use.}
As a consequence, to the best of our knowledge, no current SFM satisfies the Open Source AI definition and can hence claim to be an Open Source SFM (OSSFM).

Considering the 24 official languages of the European Union (EU),\footnote{\url{https://european-union.europa.eu/principles-countries-history/languages_en}} we take the first step towards filling this gap and, in particular, toward the creation of an EU-OSSFM: the collection of \sara{OS-}compliant training data.
With this aim, we survey the automatic speech recognition (ASR) datasets and the unlabeled speech corpora available for EU languages and list those that can be used to train an EU-OSSFM, for a total of 950k hours. This inventory of \sara{OS-compliant} data, which will be continuously updated, is called \mosels{} (\textbf{\underline{M}}assive \textbf{\underline{O}}pen-source compliant \textbf{\underline{S}}peech data for the \textbf{\underline{E}}uropean \textbf{\underline{L}}anguages) and is publicly available as a GitHub repository at: \github{} \href{https://github.com/hlt-mt/mosel}{\texttt{github.com/hlt-mt/mosel}}. In addition, to further ease the development of an EU-OSSFM, we automatically generated transcripts (i.e., pseudo-labels) for 441k hours of unlabeled data, which we release under the \sara{OS-compliant} CC-BY 4.0 license on HuggingFace at: \huggingface{}~\href{https://huggingface.co/datasets/FBK-MT/mosel}{\texttt{hf.co/datasets/FBK-MT/mosel}}. We conclude our work with an experiment on Maltese, one of the lowest-resourced 
languages, \sara{showing} that the data can effectively be used for training ASR models.

\section{Open Source Compliant Speech Data}
\label{sec:survey}

\begin{table*}[!ht]
\footnotesize
\setlength{\tabcolsep}{1.5pt}
    \centering
        \begin{tblr}{
        rows={rowsep=1.8pt},
      colspec={|X[1.3]|X[0.5,c]|X[0.325,c]|X[1.2,c]|X[0.2,c]|}, row{1} = {c}, hlines,
    }
        \textbf{Name} & \textbf{License} & \textbf{Hours} & \textbf{Languages} & \textbf{Label} \\
        \hline
        CommonVoice \citep{commonvoice:2020} & CC-0 & 6,732 & bg, cs, da, nl, en, et, fi, fr, de, el, hu, ga, it, lv, lt, mt, pl, pt, ro, sk, sl, es, sv & \yes \\
        CoVoST2 \citep{wang21s_interspeech} & CC-0 & 687 & en, fr, it, es, pt, et, nl, sv, lv, sl & \yes \\
        CSS10 \citep{park19c_interspeech} & Public Domain & 99 & nl, fi, fr, de, el, hu, es & \yes \\
        EMU \citep{11321/236} & CC-BY 3.0 & 56 & pl & \yes \\
        EU Parliament \citep{11321/821} & CC-BY 4.0 & 32 & pl & \yes \\
        FLEURS \citep{10023141} & CC-BY 4.0 & 215 & bg, cs, da, nl, en, et, fi, fr, de, el, hu, ga, it, lv, lt, mt, pl, pt, ro, sk, sl, es, sv & \yes \\
        Large Corpus of Czech Parliament Plenary Hearings \citep{kratochvil-etal-2020-large} & CC-BY 4.0 & 444 & cs & \yes \\
        LibriLight \citep{librilight} & Public Domain & 57,706 & en & \no \\
        LibriTTS \citep{zen19_interspeech} & CC-BY 4.0 & 585 & en & \yes \\
        LibriSpeech \citep{7178964} & CC-BY 4.0 & 360 & en &  \yes \\
        LibriVoxDeEn \citep{beilharz-etal-2020-librivoxdeen} & Public Domain & 547 & de & \yes \\
        MC Speech \citep{mcspeech} & CC-0 & 22 & pl & \yes \\
        MLS (Multilingual LibriSpeech) \citep{pratap20_interspeech} & CC-BY 4.0 & 50,687 & nl, en, fr, de, it, pl, pt, es & \yes \\
        SIWIS \citep{Honnet2017TheSF} & CC-BY 4.0 & 11 & fr & \yes \\
        Speech Commands \citep{warden2018speech} & CC-BY 4.0 & 18 & en & \yes \\
        VCTK \citep{Yamagishi2019CSTRVC} & CC-BY 4.0 & 44 & en & \yes \\
        {\SetCell[r=2]{l}VoxPopuli \citep{wang-etal-2021-voxpopuli}} & {\SetCell[r=2]{c}CC-0} & 383,500 & bg, hr, cs, da, nl, en, et, fi, fr, de, el, hu, it, lv, lt, mt, pl, pt, ro, sk, sl, es, sv & \no \\
        &  & 1,791 & hr, cs, nl, en, et, fu, fr, de, hu, it, lt, pl, ro, sk, sl, es & \yes \\
        {\SetCell[r=2]{l}YouTube-Commons \citep{huggingfacePleIAsYouTubeCommonsDatasets}} & {\SetCell[r=2]{c}CC-BY 4.0} & 3,261 & bg, cs, nl, en, et, fr, de, el, hu, it, pl, pt, ro, es & \no \\
        & & 443,396 & bg, cs, nl, en, et, fi, fr, de, el, hu, it, lv, lt, pl, pt, ro, es, sv & \yes \\
        \end{tblr}
    \caption{\mosels{} speech datasets with \sara{OS-compliant} license. We also report the total number of hours (\textit{Hours}), languages supported (\textit{Languages}), and whether they include reference transcripts (\textit{Label}).}
    \label{tab:open-datasets}
\end{table*}

This section surveys the available 
corpora that are admissible for developing an OSSFM for all 24 official EU languages. 
Accordingly, we include datasets that are freely accessible (i.e., excluding paid datasets) and whose data is released under an OS-compliant license \sara{(}i.e.\sara{,} without restrictions on creating and redistributing derivative artifacts, including AI models\sara{)}.\footnote{\url{https://creativecommons.org/faq/\#artificial-intelligence-and-cc-licenses}}
This means that, in the case of the widespread Creative Commons (CC) licenses, we cannot include data released with non-derivative (ND), non-commercial (NC), or share-alike (SA)\footnote{As the license of the resulting model ``\textit{must be a Creative Commons license with the same License Elements [...] or a BY-SA Compatible License}'' (\url{https://creativecommons.org/licenses/by-sa/4.0/legalcode\#s3b}), which is not compliant with open source terms.}  restrictions. We also exclude datasets whose license is OS compliant but \sara{containing} data released under a non-OS-compliant license.
In fact, CC licenses ``\textit{allow licensed material to be included in collections} [...]\textit{, however this does not change the license applicable to the original material}''.\footnote{\url{https://creativecommons.org/faq/\#if-i-create-a-collection-that-includes-a-work-offered-under-a-cc-license-which-licenses-may-i-choose-for-the-collection}}

In line with this indication, in cases where the transcripts are OS compliant (e.g., CC-BY where only attribution is required) but the corresponding speech (or part of it) is not, we document the dataset under the most restrictive license.
For instance, GigaSpeech \citep{chen21o_interspeech}, which is released under Apache 2.0,\footnote{\url{https://apache.org/licenses/LICENSE-2.0}} is categorized as non-OS compliant since it contains YouTube videos under restrictive CC licenses.\footnote{\url{https://www.youtube.com/static?template=terms}}
Similarly, MaSS \citep{zanon-boito-etal-2020-mass} and CMU Wilderness \citep{8683536} are regarded as non-OS compliant since they are derived from the \textit{Bible.is} data of the \textit{Faith Comes By Hearing} organization with NC and ND terms of use\mg{.}\footnote{\url{https://www.faithcomesbyhearing.com/terms}}

Table \ref{tab:open-datasets} lists the OS-compliant datasets with their license, number of hours, supported languages,\footnote{Represented as two-letter ISO 639 codes: \url{https://en.wikipedia.org/wiki/List_of_ISO_639_language_codes}.} and whether they also contain transcripts.\footnote{For completeness, in Appendix 
\ref{sec:nonopen}, we list the most popular non-OS-compliant datasets, divided into those licensed under SA (Table \ref{tab:sa-datasets}), and under NC, ND, and other licenses (Table \ref{tab:nonopen-datasets}).}
The resulting \mosels{} collection comprises 18 datasets, 7 of which are either in the Public Domain -- i.e., without copyright terms \citep{fishman2006public} -- or licensed under CC-0, the most permissive CC license.\footnote{\url{https://creativecommons.org/share-your-work/cclicenses/}}
Overall, there are 505,7k hours of labeled data (i.e., including the transcripts).
However, 87\% of it comes from the YouTube-Commons dataset \citep{huggingfacePleIAsYouTubeCommonsDatasets}, for which manual inspection revealed some issues, as \textit{i)} it includes videos without speech (e.g., with only music), \textit{ii)} the language identification (LID) tag and the transcripts are often inaccurate, and \textit{iii)} sentence-level segmentation of the speech is not provided (it contains unsegmented transcripts for the entire YouTube videos).
Therefore, further checks and processing work would be needed to effectively exploit the dataset for OSSFM training.

The total speech content (both labeled and unlabeled) amounts to 950,2k hours, which
significantly exceeds the total data used to train most of the current SFMs (e.g., 680k hours for Whisper v2, 180k for OWSM), with the only exception of Whisper v3 whose training data comprises 5 million of hours.
Even excluding the 446k hours of YouTube-Commons, the amount of data remains comparable, especially since Whisper v2 and OWSM target 99 and 151 languages respectively, instead of the 24 required for an EU-OSSFM.

Looking at language coverage, Table \ref{tab:dataset-hours} shows that labeled data distribution is highly skewed towards English \sara{(see also Figure \ref{fig:label-distrib} in Appendix
\ref{subsec:data-distribution})}.
Indeed, only 6 other languages (de, es, fr it, nl, pt) can be considered as high-resource, with more than 3k hours. Instead, the unlabeled data \mg{is more evenly distributed (see also Figure \ref{fig:unlabel-distrib} in 
Appendix
\ref{subsec:data-distribution})} and includes at least 8k hours for all EU languages but Irish, for which, unfortunately, we did not find unlabeled OS-compliant data.

\begin{table}[!t]
\footnotesize
    \centering
    \begin{tabular}{c|cc||c}
       \textbf{Language} & \textbf{Label.} & \textbf{Unlabel.} & \textbf{Total} \\
       \hline
        Bulgarian (bg) & 111 & 17,609 & 17,720 \\
        Croatian (hr) & 55 & 8,106 & 8,161 \\
        Czech (cs) & 591 & 18,705 & 19,296 \\
        Danish (da) & 20 & 13,600 & 13,620 \\
        Dutch (nl) & 3,395 & 19,014 & 22,409 \\
        English (en) & 437,239 & 84,704 & 521,943 \\
        Estonian (et) & 60 & 10,604 & 10,664 \\
        Finnish (fi) & 64 & 14,200 & 14,264 \\
        French (fr) & 26,984 & 22,896 & 49,880 \\
        German (de) & 9,236 & 23,228 & 32,464 \\
        Greek (el) & 35 & 17,703 & 17,738 \\
        Hungarian (hu) & 189 & 17,701 & 17,890 \\
        Irish (ga) & 17 & 0 & 17 \\
        Italian (it) & 3,756 & 21,933 & 25,689 \\
        Latvian (lv) & 173 & 13,100 & 13,273 \\
        Lithuanian (lt) & 36 & 14,400 & 14,436 \\
        Maltese (mt) & 19 & 9,100 & 9,119 \\
        Polish (pl) & 510 & 21,207 & 21,717 \\
        Portuguese (pt) & 5,492 & 17,526 & 23,018 \\
        Romanian (ro) & 121 & 17,906 & 18,021 \\
        Slovak (sk) & 61 & 12,100 & 12,161 \\
        Slovenian (sl) & 32 & 11,300 & 11,332 \\
        Spanish (es) & 17,471 & 21,526 & 38,997 \\
        Swedish (sv) & 58 & 16,300 & 16,358 \\
        \hline
        \textit{Total} & 505,725 & 444,467 & 950,192 \\
    \end{tabular}
     \caption{\mosels{} number of hours of labeled and unlabeled speech data for each official EU language.}
    \label{tab:dataset-hours}
\end{table}

\section{Pseudo-labeling Process}
\label{sec:transcription-process}

The \mg{statistics reported} \sara{in} \S\ref{sec:survey} \mg{highlight} the importance of leveraging unlabeled data for training an OSSFM, given the scarcity of labeled material for most languages.
When unlabeled data is available for model training, a common strategy consists of creating weak supervision \citep{10.1093/nsr/nwx106,8683343,oramas-etal-2021-bootstrapping,9795080,ren-2023-weakly-supervised}, which, in the context of ASR, entails generating automatic transcripts.
In light of the high computational resources demanded by this process \sara{for large-scale SFM training data,} avoiding duplicated efforts across different institutions can significantly reduce the overall environmental impact and costs \citep{strubell-etal-2019-energy}, in line with Green AI principles \citep{schwartz2019green}. 
For this reason, we complement our inventory by providing practitioners with automatic transcripts for 441k hours of unlabeled speech coming from VoxPopuli and LibriLight.\footnote{YouTube-Commons was excluded 
due to the issues described in \S\ref{sec:survey}.}
The resulting pseudo-labeled data, whose statistics per language are presented in \S\ref{subsec:pseudolab-stats}, covers nearly half of the total data available for training an EU-OSSFM
and 23 of the 24 EU languages.
In line with the spirit of this work, the transcripts are released under the OS-compliant CC-BY license at \href{https://huggingface.co/datasets/FBK-MT/mosel}{\texttt{hf.co/datasets/FBK-MT/mosel}}.

The data is transcribed using Whisper large v3\footnote{\url{https://huggingface.co/openai/whisper-large-v3} with HuggingFace v4.38.2.}, which is released under the OS Apache 2.0 License that allows the generated content to be released under any license. In Appendix \ref{app:whisper_asr}, we report the ASR quality of Whisper across the EU languages. The inference is realized by feeding Whisper with the corresponding language ID and the \texttt{<|notimestamp|>} token, with 5 as beam size.
As LibriLight, differently from VoxPopuli, contains segments longer than Whisper's maximum duration limit of 30s, we split them into chunks of up to 30s each.
To ensure reproducibility\sara{,} we will \mg{release the code under the} Apache 2.0 Licence.

\paragraph{Costs.} 
\label{subsec:costs}
We executed all the inferences on NVIDIA A100 64GB GPUs, on which we managed to fit 16 samples per batch and enabled FlashAttention \citep{dao2022flashattention} to speed up the generation process. In this way, we reached a throughput of $\sim$1.5-2k samples per GPU hour. As a result, the transcription process required a total of $\sim$25,500 GPU hours. On popular cloud services such as AWS, this would translate to $>$100k USD\footnote{As of June 10\textsuperscript{th} 2024, 8 A100 GPUs cost $>$32 USD. See \url{https://aws.amazon.com/it/ec2/instance-types/p4/}.} and 35,625 kgCO$_2$eq estimated emissions.\footnote{Estimations were conducted using the \href{https://mlco2.github.io/impact\#compute}{MachineLearning Impact calculator} presented in \cite{lacoste2019quantifying}.}

\section{Proof of Concept on Maltese}
\label{sec:proof}

\begin{table}[t]
\small
\setlength{\tabcolsep}{4pt}
\centering
\begin{tabular}{l|c|c}
\hline
\textbf{Model} &{\bf CommonVoice }& {\bf FLEURS}\\
\hline
Whisper large v3 & 80.8 & 73.8 \\
\hline
label. + pseudo-lab. & 39.4 & 38.9 \\
label. + \textit{filtered} pseudo-lab. & 23.8 & 24.5 \\
\hline
\end{tabular}
\caption{ASR results (WER$\downarrow$) for Maltese. We compare Whisper and our models trained respectively \textit{i)} on labeled and pseudo-labeled \mosels{} data and \textit{ii)} on the same data with filters applied to pseudo-labeled data.}
    \label{tab:poc_mt}
\end{table}

To showcase that the datasets collected in our survey (\S\ref{sec:survey}) and the generated transcripts (\S\ref{sec:transcription-process}) constitute suitable training data for an EU-OSSFM, we conduct a proof-of-concept experiment on Maltese.
Maltese was chosen because it is \textit{i)} one of the lowest-resourced languages, and \textit{ii)} 
the one for which Whisper achieves the worst results, as shown in Appendix \ref{app:whisper_asr}.\footnote{With the only exception of Irish, which has only 17 collected hours and is not even supported by Whisper.}

\lb{For our experiments, we first attempted to train an ASR model using only supervised data, but it failed to converge due to its limited size (16 hours). Therefore, we trained a model using the few labeled data together with the pseudo-labeled data.\footnote{For full experimental details see Appendix
\ref{sec:exp-sett}.}  As an additional investigation, we also applied to the pseudo-labeled data simple filtering methods to remove audios containing other languages and automatic transcripts containing hallucinations (see Appendix
\ref{sec:data-filtering}).
 Results presented in Table \ref{tab:poc_mt} show that the model trained with all data doubles the performance of Whisper ($\sim$39 vs. $\sim$80 WER). Considering the very low performance of Whisper, which was used to create the automatic transcripts, the contribution of the pseudo-labeled data is noticeable. 
Also interesting is the further improvement obtained when unlabeled data are filtered ($\sim$24 WER).
These experiments support the conclusion that the collected and transcribed data represent a promising bedrock for developing an EU-OSSFM.}

\section{Conclusions}

In response to the urgent need for truly open-source foundation models, this work takes the first step toward an EU open-source speech foundation model, which is the collection of suitable training data called \mosels{}. 
To this end, we first surveyed the labeled and unlabeled speech datasets for automatic speech recognition that feature at least one of the 24 official EU languages and are available under a license compliant with the open-source terms.
We then complemented this effort with the creation and release of automatic transcripts for the available unlabeled data.
Overall, we collected more than 950k hours of speech content suitable for the training of an EU open-source speech foundation model, also demonstrating its usefulness in Maltese, one of the lowest-resourced languages.

\section{Acknowledgments}
The work presented in this paper is funded by the European Union's Horizon research and innovation programme under grant agreement No 101135798, project Meetween (My Personal AI Mediator for Virtual MEETtings BetWEEN People) and the PNRR project FAIR -  Future AI Research (PE00000013),  under the NRRP MUR program funded by the NextGenerationEU. We also acknowledge the CINECA award (MAGIS) under the ISCRA initiative, for the availability of high-performance computing resources and support.

\section{Limitations}


\paragraph{Collecting Open Irish Data.}
\mg{An important future direction to expand this work}
is represented by collecting and releasing new material 
-- possibly 
with human-generated transcripts --
under permissive licenses for the least-resourced language. This is especially critical for Irish, for which we were able to collect only 17 hours of (labeled) speech.

\paragraph{Data Curation of Available Resources.}
As noted in \S\ref{sec:survey}, the quality of the supervision of the surveyed dataset cannot always be taken for granted, advocating for dedicated inspections before using it to train an OSSFM. 
This is particularly true for the metadata and transcripts of YouTube videos \sara{under OS-compliant licenses} as those collected in YouTube-Commons.

\paragraph{Quality of Pseudo-labels and Filtering Techniques.}
The quality of Whisper outputs greatly varies across the 24 languages. In particular, the WER of Whisper for Maltese is high (80.8 on the CommonVoice test set and 73.8 on the FLEURS test set). As such, filtering strategies aiming at identifying unreliable transcriptions may be required for the successful training of OSSFM, especially for low-resource languages. \sara{Indeed, as already seen in \S\ref{sec:proof}, even simple filtering techniques proved to be effective in greatly improving ASR performance.} \mg{More advanced filtering techniques can provide further benefits for the quality of the resulting model.} However, data cleaning and normalization are common steps in training pipelines, going beyond the scope of this work.

\paragraph{Beyond EU languages.} This paper has focused only on the 24 EU languages. An obvious next step for this work is its extension to 
many other spoken languages, with the final goal of covering hundreds of languages \mg{and leading to the creation of a universal OSSFM}.



\bibliography{custom}

\newpage
\appendix

\section{Data Statistics}

\subsection{Labeled and Unlabeled Data Distribution}
\label{subsec:data-distribution}

\begin{figure}[h]
    \centering
    \begin{subfigure}[hb]{0.475\textwidth}
            \centering
            \includegraphics[width=\linewidth, 
            ]{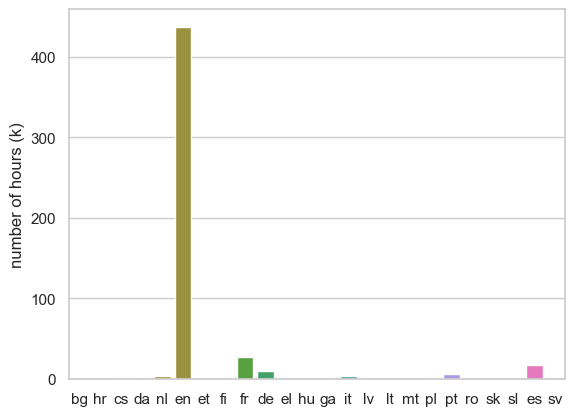}
            \caption{Labeled}
            \label{fig:label-distrib}
        \end{subfigure}\\
    \begin{subfigure}[hb]{0.475\textwidth}
            \centering
            \includegraphics[width=\linewidth, 
            ]{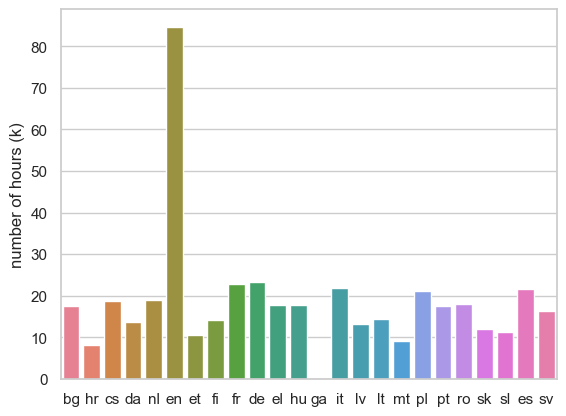}
            \caption{Unlabeled}
            \label{fig:unlabel-distrib}
        \end{subfigure}
    \caption{Labeled and unlabeled data distribution of the OS-compliant collected speech for each EU language.}
    \label{fig:label-unlabel-distrib}
\end{figure}

Data distributions for both labeled and unlabeled data discussed in \S\ref{sec:survey} and referred to Table \ref{tab:open-datasets} are presented in, respectively, Figure \ref{fig:label-distrib} and \ref{fig:unlabel-distrib}.

\subsection{Pseudo-labeled Data Statistics}
\label{subsec:pseudolab-stats}

The total number of hours of pseudo-labeled data described in \S\ref{sec:transcription-process} are shown in Table \ref{tab:pseudolabel-hours}. The data distribution is similar to those of unlabeled data presented in \S\ref{subsec:pseudolab-stats} due to the nearly complete overlap with the retrieved unlabeled data, as already discussed in \S\ref{sec:transcription-process}.

\begin{table}[!t]
\footnotesize
    \centering
    \begin{tabular}{c|cc||c}
       \textbf{Language} & \textbf{Pseudo-labeled  (hours)} \\
       \hline
        Bulgarian (bg) & 17,600 \\
        Croatian (hr) & 8,100 \\
        Czech (cs) & 18,700 \\
        Danish (da) & 13,600 \\
        Dutch (nl) & 19,000 \\
        English (en) & 81,806 \\
        Estonian (et) & 10,600 \\
        Finnish (fi) & 14,200 \\
        French (fr) & 22,800 \\
        German (de) & 23,200 \\
        Greek (el) & 17,700 \\
        Hungarian (hu) & 17,700 \\
        Italian (it) & 21,900 \\
        Latvian (lv) & 13,100 \\
        Lithuanian (lt) & 14,400 \\
        Maltese (mt) & 9,100 \\
        Polish (pl) & 21,200 \\
        Portuguese (pt) & 17,500 \\
        Romanian (ro) & 17,900 \\
        Slovak (sk) & 12,100 \\
        Slovenian (sl) & 11,300 \\
        Spanish (es) & 21,400 \\
        Swedish (sv) & 16,300 \\
        \hline
        \textit{Total} & 441,206 \\
    \end{tabular}
     \caption{Number of hours for the pseudo-labeled data that we make available for each official EU language.}
    \label{tab:pseudolabel-hours}
\end{table}

\section{Experimental Settings}
\label{sec:exp-sett}

\subsection{Model and Training Settings}
\label{sec:model-training-sett}

We train a sequence-to-sequence model whose encoder is a 12-layer Conformer \citep{gulati20_interspeech} and whose decoder is a 6-layer Transformer \citep{transformer}. The Conformer encoder is preceded by two 1D convolutional layers with stride 2 and kernel size 5. We use an embedding size of 512 and an internal feed-forward dimension of 2048. The convolutional modules of the Conformer layers have a 31-feature kernel. The target vocabulary is built with size 8,000 using SentencePiece \citep{kudo-2018-subword}, while the input audio is represented with 80 Mel-filterbank features extracted every 10 ms with a window of 25 ms.
As a result, the model has 116M parameters in total.

We use label-smoothed cross-entropy loss on the decoder output (with 0.1 as label-smoothing factor), complemented with a CTC \citep{ctc-2006} loss (summed with 0.5 weight) trained on the output of the 8th encoder layers to facilitate the convergence of the model. The model was optimized with Adam ($\beta_1,\beta_2=0.9,0.98$) using Noam learning rate scheduler \citep{transformer} with 2e-3 as peak learning rate and 25,000 warmup steps. To avoid overfitting, we set dropout to 0.1 and weight decay to 0.001 and apply SpecAugment \cite{Park2019} during training.
To further ease the convergence of the model, we initialize the Conformer encoder weights with those of a similar ASR model trained on 4k hours of labeled English data, comprising CommonVoice, Librispeech, CoVoST, and VoxPopuli.
We train the models with mini-batches of 40,0000 tokens and 2 as update frequency on 4 NVIDIA Ampere A100 GPUs (64GB RAM) for 150k updates and average the last 7 checkpoints.

Our experiments are conducted with the open-source
\mmg{repository available at \url{https://github.com/hlt-mt/FBK-fairseq/} using the padding-safe implementation of the Conformer encoder \citep{papi-etal-2024-good}}.
\sara{Results in Word Error Rate (WER) are computed using the Whisper Normalizer\footnote{\url{https://pypi.org/project/whisper-normalizer/}} and, then, JiWER\footnote{\url{https://pypi.org/project/jiwer/}} for computing the metric.}

\subsection{Data Filtering}
\label{sec:data-filtering}

\subsubsection{LID}


To check for possible inconsistencies between the metadata released in VoxPopuli and the actual content of speech segments, we check the actual spoken language with an automatic language identifier (LID). In fact, as in the transcription process described in \S\ref{sec:transcription-process} we force the language to the one provided in the metadata, these segments may be paired with noisy transcripts. The LID was carried out using the Whisper {\it large v3} model, as done for the transcription process, and it was performed by letting the model predict the language tag and taking the language with the highest probability.

\begin{table}[ht]
\small
\centering
\begin{tabular}{ c | c  }
{\bf LID } & {\bf Portion (\%)}  \\ 
\hline
mt   &  77.1 \\
en & 9.9  \\
it & 3.5 \\
fr & 2.2 \\
ar & 1.9  \\
other & 5.4 \\
\end{tabular}
\caption{Identified languages on the Maltese section of VoxPopuli (reported as \%).}
\label{tab:lid}
\end{table}

\begin{table*}[!ht]
\footnotesize
    \centering
    \begin{tabular}{c|l|l}
\hline
\textbf{id} & \textbf{Reference} & \textbf{Automatic Transcript} \\
\hline
1 & Where is the victim? & Yes, where's the victim? \\
2 & Here & \textcolor{red}{Hey, hey, hey,} here, \textcolor{red}{hey}. \textcolor{red}{No, no, no, no, no, no, no, no.}\\
3 & Good, and now to get hold of little Henny.
& Shop for a moment, give Hennie a hand. \\
\hline
\end{tabular}
\caption{Examples of hallucinations in Whisper outputs.}
\label{tab:hall}
\end{table*}

Table \ref{tab:lid} shows the results. Upon a manual inspection, we noticed that the samples predicted as Maltese are indeed all correct. Similarly, the LID appeared mostly correct when predicting languages different from Maltese, except for the samples identified as Italian or Arabic which are sometimes Maltese speech. However, given the not-so-high amount of mislabeled data and to be on the safe side, in our experiments with ``simple filters'' we opted for filtering all the data recognized with a language different from Maltese, removing $\sim$23\% of the 9k VoxPopuli hours.

To ensure the reproducibility of our experiments and to let practitioners leverage this information for their filtering strategies while creating OSSFM, we will release the LID output for all the transcribed unlabeled data under the CC-BY license.

\subsubsection{Textual Hallucinations}

In the context of LLMs, hallucinations refer to ``\textit{the generation of content that deviates from the real facts, resulting in unfaithful outputs}'' \citep{maynez-etal-2020-faithfulness,rawte-etal-2023-troubling}. In our context of ASR, they have been analogously defined as ``\textit{nonsensical, or unfaithful to the provided source input}'' \citep{10.1145/3571730}.
 %
Specifically, here we focus on the detection of \textit{nonsensical} hallucinations, in which ``\textit{the generated text fails to convey any relevant or comprehensible information}'',\footnote{\scriptsize\url{https://masterofcode.com/blog/hallucinations-in-llms-what-you-need-to-know-before-integration}}
 while those related to semantic aspects are ignored.
 
 Table~\ref{tab:hall} shows 
 examples of hallucinated 
 texts generated by Whisper
 in English.
It can be noted that, in 
line 2, the word \textit{here} is surrounded by many spurious ``\textit{hey,}''
and that the successive 
sentence consists of a sequence of equally spurious ``\textit{no,}''.
This typically happens when 
background noise or music is present in the audio content, making the transcription task more difficult.

Another issue that can affect, although less frequently, the text generated by LLMs in general and by Whisper in particular, is the presence of very long and noisy strings like \enquote{T-J-N-D-F-Z-3-2-8-W-M-L-G-0-Z-P-[$\ldots$]} and \enquote{Amen.Amen.Amen.Amen.Amen.Amen.[$\ldots$]}.
Moreover, we noted that\sara{,} for some languages\sara{,} 
the decoding of entire audio 
segments
\sara{sometimes generates}
one single, very common word, like ``\textit{Děkuji}'' for Czech and ``\textit{Ačiū}'' for Lithuanian, both corresponding to ``\textit{Thank you}''. 
\sara{Although being correct in some cases,}
since for the most reliable languages (e.g., English and German) transcripts with a single word are relatively rare, we 
chose to 
\sara{consider} this phenomenon 
\sara{as hallucination.}

In conclusion, we 
decided to flag the segments containing all the above-described 
\sara{hallucinations,}
with the option of filtering them out during training. Also in this case, for the sake of 
reproducibility
and 
\sara{to enable} the adoption of similar approaches, we released the hallucination-detection metadata.

\begin{table}[t]
\centering
\small
\begin{tabular}{c|cc}
 \textbf{Language}  & \textbf{CommonVoice} & \textbf{FLEURS}    \\
   \hline
bg & 14.3        & 12.5      \\
hr & -           & 10.8      \\
cs & 9.0        & 10.1      \\
da & 18.1        & 12.0     \\
nl & 4.3         & 5.2       \\
en & 9.3         & 4.1       \\
et & 29.9        & 18.1      \\
fi & 24.6        & 7.7       \\
fr & 10.8        & 5.3       \\
de & 5.7         & 4.9       \\
el & 13.7        & 10.9      \\
hu & 13.4        & 12.9      \\
ga & -        & -      \\
it & 5.5         & 3.0       \\
lv & 16.7        & 19.4      \\
lt & 27.6        & 23.7      \\
mt & 80.8         & 73.8 \\
pl & 6.0         & 4.6       \\
pt & 5.9         & 4.1       \\
ro & 10.8        & 8.2      \\
sk & 23.4        & 9.2      \\
sl & 16.8        & 18.3      \\
es & 4.7         & 2.8       \\
sv & 8.3        & 7.6   
\end{tabular}
\caption{WER ($\downarrow$) reported for Whisper large v3 \citep{radford2019language} across the 24 European languages on CommonVoice and FLEURS. }
\label{tab:whisper_wer}
\end{table}


\section{Non-open Datasets}
\label{sec:nonopen}

\subsection{CC-BY-SA}
\label{subsec:share-alike}

The collection of datasets with the SA license, which is not compliant with open-source criteria, is presented in Table \ref{tab:sa-datasets}.

\subsection{CC-NC, -ND, and others}
\label{subsec:nc-nd-others}

The collection of the most well-known datasets with a license that is not compliant with open-source criteria
is presented in Table \ref{tab:nonopen-datasets}.

\begin{table*}[!ht]
\footnotesize
\setlength{\tabcolsep}{1pt}
    \centering
        \begin{tblr}{
      colspec={|X[1.5]|X[0.5,c]|X[0.325,c]|X[1,c]|X[0.2,c]|}, row{1} = {c}, hlines,
    }
        \textbf{Name} & \textbf{License} & \textbf{hours} & \textbf{Languages} & \textbf{Label} \\
        \hline
        ARTHUR 1.0 \citep{11356/1772} & CC-BY-SA 4.0 & 884 & sl & \yes \\
        Vystadial \citep{korvas-etal-2014-free} & CC-BY-SA 3.0 & 63 & en, cs & \yes\\
        ParlaSpeech-HR \citep{ljubesic-etal-2022-parlaspeech} & CC-BY-SA & 1,816 & hr & \yes \\
        People's Speech \citep{galvez2021people} & CC-BY-SA 4.0 & 30,000 & en & \yes \\
        SWC \citep{KHN16.518} & CC-BY-SA 4.0 & 996 & de, en, nl & \no \\
        SWC-ASR \citep{KHN16.518} & CC-BY-SA 4.0 & 510 & de, en, nl & \yes \\
        UK and Ireland English Dialect \citep{demirsahin-etal-2020-open} & CC-BY-SA 4.0 & 31 & ga & \yes\\
        \end{tblr}
    \caption{Speech datasets with Share-Alike (SA) license. If more languages are included, the sum is presented.}
    \label{tab:sa-datasets}
\end{table*}

\begin{table*}[!ht]
\footnotesize
\setlength{\tabcolsep}{0.5pt}
    \centering
        \begin{tblr}{
      colspec={|X[1.5]|X[0.8,c]|X[0.4,c]|X[c]|X[0.25,c]|}, row{1} = {c}, hlines,
    }
        \textbf{Name} & \textbf{License} & \textbf{hours (k)} & \textbf{Languages} & \textbf{Label} \\
        \hline
        AMI \citep{10.1007/11677482_3} & CC-BY-NC 4.0 & 100 & en & \yes \\
        AudioCite.net \citep{Felice2024} & CC-BY-NC & 6,682 & fr & \yes \\
        BEA-Base \citep{mihajlik-etal-2022-bea} & NC & 71 & hu & \yes \\ 
        CMU Wilderness \citep{8683536} & NC, ND & 236 & en, fi, fr, pl, pt, to, es, sv & \yes \\
        Europarl-ST \citep{jairsan2020a} & CC-BY-NC 4.0 & 201 & en, fr, de, it, es, pt, pl, ro, nl & \yes\\
        FT Speech \citep{ftspeech} & NC & 1,800 & da & \yes \\
        GigaSpeech \citep{chen21o_interspeech} & YouTube License & 33,000 & en & \no \\
        GigaSpeech-ASR \citep{chen21o_interspeech} & YouTube License & 10,000 & en & \yes \\
        GOS \citep{gos} & CC-BY-SA-NC 2.5 & 120 & sl & \yes \\
        How-2 \citep{sanabria18how2} & YouTube License & 2,000 & en & \no \\
        How-2 ASR \citep{sanabria18how2} & YouTube License & 300 & en & \yes \\
        M-AILABS \citep{caitoMAILABSSpeech} & Project Gutenberg License & 867 & en, fr, de, it, pl, es & \yes  \\
        MASRI \citep{hernandez-mena-etal-2020-masri} & NC & 8 & mt & \yes \\
        MaSS \citep{zanon-boito-etal-2020-mass} & NC, ND & 126 & en, fi, fr, hu, ro, es & \yes \\
        MediaSpeech \citep{mediaspeech2021} & YouTube License & 20 & fr & \yes \\
        mTEDx \citep{salesky2021mtedx} & CC-BY-NC-ND 4.0 & 679 & fr, el, it, pt, es & \yes \\
        MuAViC \citep{anwar2023muavic} & CC-BY-NC 4.0 & 1,079 & en, el, es, fr, it, pt & \yes \\
        MuST-C \citep{di-gangi-etal-2019-must} & CC-BY-NC-ND 4.0 & 504 & en & \yes \\
        PDTSC1.0 \citep{11234/1-2375}  & CC-BY-NC-SA 4.0 & 122 & cs & \yes\\
        PELCRA \citep{pezik-2018-increasing} & CC-BY-NC & 100 & pl & \yes \\
        SpokesBiz \citep{pkezik2023spokesbiz} & CC-BY-NC-ND & 650 & pl & \yes \\
        SPGISpeech \citep{oneill21_interspeech} & NC & 5,000 & en & \yes \\
        SWARA \citep{stan-sped2017} & CC-BY-NC 4.0 & 21 & ro & \yes \\ 
        Tatoeba ENG \citep{tatoebaIndexaudio} & CC-BY-NC-ND & 200 & en & \yes \\
        TEDLIUM v3 \citep{hernandez2018ted} & CC-BY-NC-ND 3.0 & 452 & en & \yes \\
        TEDx Spanish \citep{mena_2019} & CC-BY-NC-ND 4.0 & 24 & es & \yes \\
        VoxLingua107 \citep{valk2021slt} & YouTube License & 1,352 & bg, hr, cs, da, nl, en, et, fi, fr, de, el, hu, it, lv, lt, mt, pl, pt, ro, sk, sl, es, sv & \no \\
        \end{tblr}
    \caption{Non-open speech datasets. If more languages are included, the sum is presented.}
    \label{tab:nonopen-datasets}
\end{table*}

\section{Whisper Performance on EU Languages}
\label{app:whisper_asr}
Table \ref{tab:whisper_wer} reports the WER scores obtained using Whisper on the 24 European languages. Maltese stands out as the worst language by a wide margin, with a very high WER (73.8 on FLEURS) indicating a limited ability to address the Maltese ASR task. All other languages display much lower WER, as only Estonian, Latvian, Lithuanian, and Slovenian exceed 15 WER, while high-resource languages such as Dutch, English, Italian, German, and Spanish consistently achieve WER close or lower than 5.

\end{document}